\documentclass[11pt]{article} 
\usepackage{rldmsubmit,palatino}
\usepackage{graphicx}
\usepackage{subcaption}
\usepackage{amsmath}
\usepackage{amssymb}

\usepackage{titlesec}
\titlespacing\section{0pt}{12pt plus 2pt minus 2pt}{0pt plus 2pt minus 2pt}
\titlespacing\subsection{0pt}{12pt plus 2pt minus 2pt}{0pt plus 2pt minus 2pt}
\title{State Space Decomposition and Subgoal Creation for Transfer in Deep Reinforcement Learning}

\author{
Himanshu Sahni\thanks{ Denotes equal contribution. } \\
College of Computing \\
Georgia Institute of Technology \\
Atlanta, GA 30332 \\
\texttt{himanshu@gatech.edu} \\
\And
Saurabh Kumar* \\
College of Computing \\
Georgia Institute of Technology \\
Atlanta, GA 30332 \\
\texttt{skumar311@gatech.edu} \\
\AND
Farhan Tejani \\
College of Computing \\
Georgia Institute of Technology \\
Atlanta, GA 30332 \\
\texttt{farhantejani@gatech.edu} \\
\And
Yannick Schroecker \\
College of Computing \\
Georgia Institute of Technology \\
Atlanta, GA 30332 \\
\texttt{yannickschroecker@gatech.edu} \\
\And
Charles Isbell \\
College of Computing \\
Georgia Institute of Technology \\
Atlanta, GA 30332 \\
\texttt{isbell@cc.gatech.edu} \\
}

\begin{document}

\maketitle

\begin{abstract}
Typical reinforcement learning (RL) agents learn to complete tasks specified by reward functions tailored to their domain. As such, the policies they learn do not generalize even to similar domains. To address this issue, we develop a framework through which a deep RL agent learns to generalize policies from smaller, simpler domains to more complex ones using a recurrent attention mechanism. The task is presented to the agent as an image and an instruction specifying the goal. This meta-controller guides the agent towards its goal by designing a sequence of smaller subtasks on the part of the state space within the attention, effectively decomposing it. As a baseline, we consider a setup without attention as well. Our experiments show that the meta-controller learns to create subgoals within the attention.
\end{abstract}



\startmain 

\section{Introduction}
Usually, reinforcement learning (RL) agents cannot generalize policies learned in small domains to larger, more complicated ones. Yet this ability can allow them to immediately apply skills learned in simple settings without having to explore a large state space. In the deep learning setting, reduction in the size of the state also means smaller networks are required, which are easier to train. In this work, we present an approach to decompose complicated environments into simpler ones and provide subgoals within them that ultimately solve the larger task. The agent can be pre-trained on the smaller environment to solve each subgoal independently, or in conjunction with the subgoal creation algorithm.

We describe a meta-controller that learns to decompose the state space and provide subgoals solvable within the smaller space. The meta-controller is solving a delayed reward problem as it only gets positive reinforcement when the underlying agent solves the original task. It has to come up with a sequence of subgoals which maximizes the expectation of this reinforcement. In addition to creating subgoals, the meta-controller also fragments the state space such that the underlying agent is presented with a smaller state on which it can easily learn an optimal policy for the subgoal. It does this by using an attention mechanism, similar to the Recurrent Attention Model \cite{DBLP:journals/corr/MnihHGK14}. The meta-controller learns to control its attention and only passes the part of the state within it to the agent. The meta-controller's MDP formulation is:
\begin{itemize}
    \item States, $S$, are summaries of its past and current attentions. 
    \item Actions, $A$, are the locations of the attention $L^{attn}$, and a distribution over the set of subgoals, $\textbf{g}$.
    \item Rewards, $r$, are positive if the underlying agent solves the task and a small negative step cost otherwise
    \item Transitions: The underlying agent executes its policy according to the state and subgoal provided to it. Since this policy is unknown to the meta-controller this is a source of stochasticity in its environment.
\end{itemize}

The meta-controller selects a value for $L^{attn}$ and a distribution $P(g)$. The state space under that location and a subgoal, $g$, are passed to the underlying agent. The agent then chooses an atomic action that moves it towards achieving $g$. The new agent location $L^{agent}$ changes the meta-controller's environment, which picks a new attention and subgoal.

In this work, we make a few simplifying assumptions. First, we assume that the underlying agent has access to the optimal policy for each subgoal. Such a goal-dependent policy can be learned by a technique such as universal value function approximators (UVFAs) \cite{Schaul:2015:UVF:3045118.3045258}. UVFAs learn to approximate $V(S,g)$, or the value function with respect to a goal, using a function approximator such as deep neural nets. The learned value function $V(S,g;\theta)$ can be used to construct a policy that achieves the goal $g$. This value function can be trained independent of or in conjunction with the meta-controller by providing intrinsic rewards for achieving subgoals \cite{DBLP:journals/corr/KulkarniNST16}. Secondly, we assume that the agent remains still unless both its location and the subgoal are present within the state provided to it by the meta-controller. In general, the meta-controller will automatically be incentivized to focus its attention and provide subgoals such that the underlying agent is able to solve the given task because of its reward structure. In this case, that means keeping both the agent location and the subgoal within the attention. For example, in the game of Pacman, if the subgoal is to eat the closest pill, the underlying agent should have both the Pacman and at least one pill within the state provided to it. Otherwise, the agent may move randomly, and it will be unable to achieve its overall goal of getting a high score.

The above assumptions simplify training of the meta-controller, but the methodology we provide should be applicable in the general setting where the policy of the underlying agent is learned as well.

\section{Related Work}
Our work most closely matches that of Kulkarni et al. \cite{DBLP:journals/corr/KulkarniNST16}. They present a hierarchical framework where an agent learns from intrinsic rewards provided by a higher level agent setting subgoals and operating on a longer time frame. Rewards for the higher level agent are provided by the environment for completing tasks. The subgoals in turn are provided through functions over entities and relations in an object oriented framework. In a sense, our approach takes this a step further by decomposing the state space such that the base agent has to only see a small portion of it at any time. This allows for better computational efficiency, as the base agent can now use smaller networks, and may allow for transfer of learnt policies to different parts of the state space that are similar without having to explicitly explore them. To achieve this, the higher level agent, or meta-controller, must learn to integrate information over the states it has observed so far. Therefore, we use a recurrent model to represent the meta-controller through a Long Short Term Memory (LSTM) network \cite{Hochreiter:1997:LSM:1246443.1246450}. Kulkarni et al. use a pair of DQNs for the agent and meta-controller. 

In order to train the attention mechanism of our meta-controller, we employ a technique similar to that of Mnih et al. \cite{DBLP:journals/corr/MnihHGK14}. They use policy gradients to train an attention mechanism for classification and simple control tasks. In our approach, we do not employ a complex glimpse sensor, but instead simply use a $5$x$5$ crop of our input image. This can be incorporated into our setup easily. Further, instead of specifying $L^{attn}$ directly using a continuous output, we use discrete actions, $up, down, noop$, to move the attention.

Finally, Schaul et al. \cite{Schaul:2015:UVF:3045118.3045258} describe how goal specific function approximators may be constructed for deep RL agents. Such a function can be constructed for our base learner by independently learning subgoals on a $5$x$5$ image. We do not provide results for this setting in this paper but it can be integrated into future work.

\section{Preliminaries}
Before proceeding further, we provide a brief introduction to reinforcement learning and policy gradients.
\subsection{Reinforcement Learning and MDPs}
Reinforcement learning addresses the problem of choosing behavior that maximizes some notion of long term cumulative reward. It is typically formulated as a Markov decision process (MDP). An MDP is characterized by the tuple $<S,A,T,R,\gamma>$. $S$ is the set of states an agent can be in. $A(S)$ is the set of actions the agent has available to it in each state. Typically, the agent chooses an action to execute from $A$, which may lead it into a new state according to the transition function $T:S \times A \rightarrow P(S)$. $R:S \times A \rightarrow \mathbb{R}$ is a scalar value received upon executing an action in a state. Finally, $0 \leq \gamma \leq 1$ is the discount factor. A policy, $\pi: S \rightarrow P(A)$, informs the agent on which action to execute in each state. The goal of a reinforcement learning agent is to find $\pi^{*}$, the optimal policy, which maximizes the long term expected reward, or utility, in each state.

\subsection{Policy Gradients}
Policy gradient methods look to directly directly maximize expected reward by adjusting policy parameters. The expected reward for a trajectory sampled from a policy $\pi$ parametrized by $\theta$ is given by

\begin{equation}
    J(\theta) = \mathbb{E}_{p(S_{1:T};\theta)} \sum_{t=1}^{T} r_{t} = \mathbb{E}_{p(S_{1:T};\theta)} [R]
\end{equation}

where, $p(S_{1:T};\theta)$ depends on the state distribution induced by $\pi$, and $R$ is the return for the trajectory. In this formulation, no discounting is performed. The policy gradient theorem \cite{Sutton:1998:IRL:551283} states that the gradient for $J$ is given by
\begin{equation}
    \nabla_{\theta} J(\theta) = \mathbb{E}_{p(S_{1:T};\theta)} \sum_{t=1}^{T} \nabla_{\theta} log[\pi(a_{t}|s_{1:t};\theta)] R_{t}
\end{equation}

The expectation can be approximated by sampling a batch of trajectories from the current policy and averaging the gradients over them. This is the REINFORCE algorithm \cite{Williams1992}. The vanilla policy gradient may suffer from high variance as it relies on Monte Carlo samples. A common modification to reduce variance is to subtract a baseline from the returns. The baseline $b_{t}$ is computed by taking the average of observed returns over the past ${N}$ time steps: $\frac{1}{N}\sum_{n=T-N+1}^{T} r_{t_{n}}$. In this work, we chose $N$ = $100$. 


\section{Environment}
For our experiments, we use an environment consisting of a $10$x$5$ grid. The grid consists of four "rooms," where each room is a horizontal $k$x$5$ strip for some $k$ not exceeding $4$. The rooms are stacked on top of each other and are each a different color from the set $\{red, green, blue, yellow\}$. The environment also generates an instruction as a one-hot vector of length $4$, specifying the target room. An episode terminates when either the agent reaches the target room, receiving a positive reward of $+1$, or it times out without reaching it. The step cost is $-0.01$.

\section{Approach}

We construct three frameworks for the meta-controller that is tasked with providing subgoals to the underlying agent such that it navigates successfully to the target room. In all experiments, the meta-controller uses the Adam Optimizer with a learning rate of $1 \times e^{-5}$. The agent and attention (if used) always start of at the top left corner of the grid.

\subsection{Meta-Controller with No Attention Mechanism}
First, we simplify the problem by providing the entire state space as input to the meta-controller at each time step, therefore not employing any attention mechanism. Specifically, the meta-controller receives as input the $10$x$5$ image of the grid which contains the rooms as well as the location of the agent, $L^{agent}$. The output of the meta-controller is a distribution $P(g)$ over the rooms. A room is sampled from $P(g)$ and is provided to the base agent as an instruction. This is the subgoal it must achieve. It is assumed that the underlying agent can move optimally on the entire $10$x$5$ grid given an instruction. In this setup, the optimal policy for the meta-controller is to always output the target room directly. 


\begin{figure}[htb]
\centering
    \begin{subfigure}[htb]{0.5\textwidth}
    \centering
    \captionsetup{justification=centering}
    \includegraphics[width=1\textwidth]{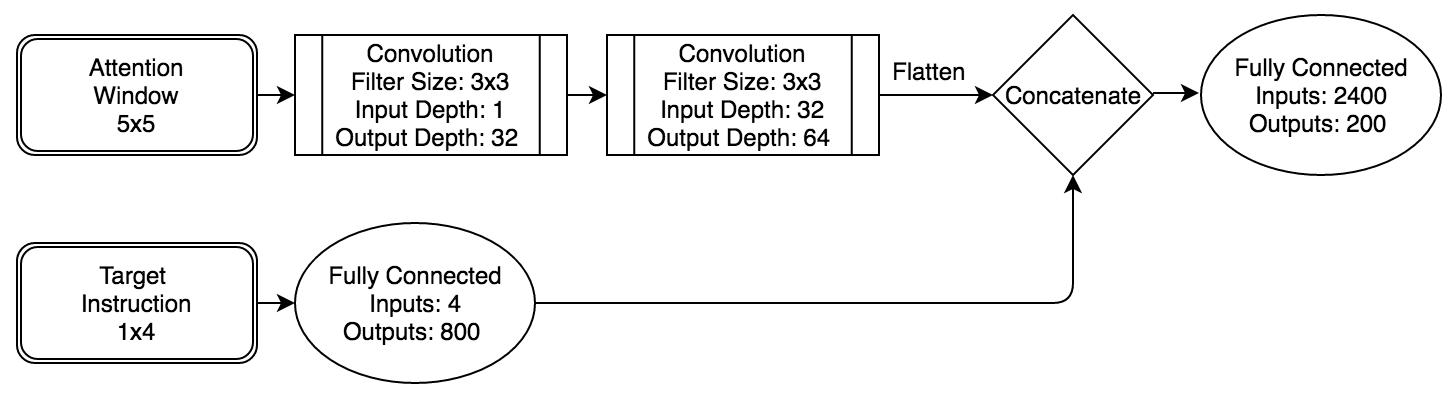}
    \captionsetup{width=0.95\textwidth}
    \caption{State Processor Network}
    \label{fig:network}
    \end{subfigure}
    \begin{subfigure}[htb]{0.3\textwidth}
    \centering
    \captionsetup{justification=centering}
    \includegraphics[width=0.99\textwidth]{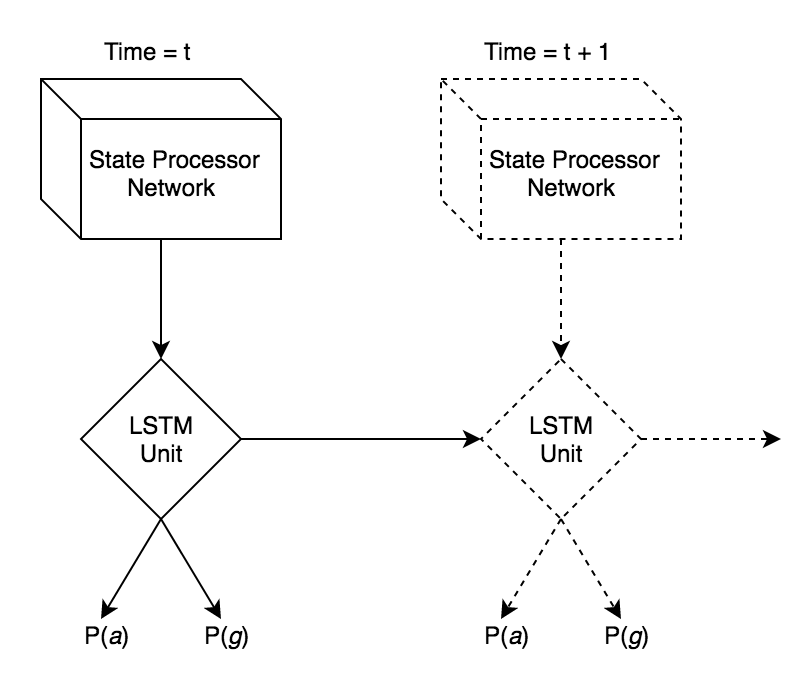}
    \captionsetup{width=0.95\textwidth}
    \caption{Architecture of Meta-Controller with Attention}
    \label{fig:architecture}
    \end{subfigure}
    \caption{Network architecture for the partial state decomposition and constrained attention mechanism experiments. The $5$x$5$ attention window and target instruction are inputted at each time step into the meta-controller, which then outputs probability distributions over attention actions and subgoal instructions.}
\end{figure}

\subsection{Meta-Controller with Partial State Decomposition}

In this setup, we have a meta-controller with an attention mechanism, which consists of a $5$x$5$ window into the $10$x$5$ grid. In addition to the subgoal instruction, it must now also output an action to control the attention. Here, the attention mechanism only partially decomposes the state space, meaning that the agent may move optimally to a provided subgoal, even if the agent is not within the current attention. The subgoal, however, must be located inside the attention. The goal of the meta-controller is to to find the location of the target room using its attention mechanism and then instruct the agent to go to that room color at every time step.

The architecture of this meta-controller consists of a state processor network, which takes the $5$x$5$ attention window and target instruction provided by the environment as input at each time step. It processes these inputs using a feedforward convolutional network and uses an LSTM unit to output $P(g)$ and $P(a)$, where $P(a)$ is the probability distribution over attention actions. The convolutional layers of the network use rectified linear unit activation functions. The attention action affects the next attention location, $L^{attn}$, while the subgoal instruction affects the next agent location, $L^{agent}$. Thus, the LSTM's hidden state contains knowledge gained from taking a sequence of both instruction and attention actions in an episode. In order to train this network effectively with Policy Gradients, we assume that the attention and instruction actions probabilities are independent of each other.

\subsection{Meta-Controller with Constrained Attention Mechanism}

In this setup the agent will not move unless it is within the attention. This means that the meta-controller must instruct the agent to move to a room within the view of the attention before moving downwards and repeating the process until the agent has reached the target room. Thus, if the agent or the subgoal do not appear within the decomposed state space, the target task will not be achieved. In both the partial state decomposition and constrained attention mechanism experiments, the LSTM unit allows the meta-controller to use its memory of the locations of the agent and target room to guide its action selection when either the agent or target room is not present within the attention window at a particular time step. The constrained attention mechnism framework adds the additional step of constructing an optimal sequence of subgoals for the agent to reach the target room, and this is the overall goal of this paper. 

\section{Results}

\subsection{Meta-Controller with No Attention Mechanism}

We ran two experiments using the meta-controller with no attention mechanism. In the first experiment, the environment is fixed, i.e. the room arrangement is fixed between episodes, and the target room is always at the very bottom. The optimal policy of the meta-controller is simply to output the instruction corresponding to the target room, since the underlying agent is an optimal agent on the entire grid. In the second experiment, the environment is dynamic, which means that the room arrangement is randomly generated between episodes, but the target room is always located at the very bottom. Here, it must learn a mapping between the color of the bottom-most room and the optimal instruction. These experiments serve as a baseline for the experiments that we run using the meta-controller with the attention mechanism. For both cases, the meta-controller converges to the optimal policy, guiding the agent to the correct room. 

\begin{figure}[htb]
\centering
    \begin{subfigure}[htb]{0.245\textwidth}
    \centering
    \captionsetup{justification=centering}
    \includegraphics[width=0.99\textwidth]{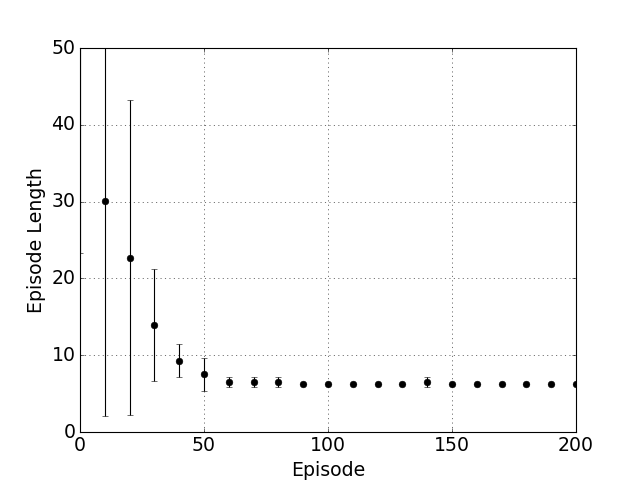}
    \captionsetup{width=0.95\textwidth}
    \caption{Meta-Controller with No Attention Mechanism on a Fixed Environment}
    \label{fig:no-attention-fixed}
    \end{subfigure}
    \begin{subfigure}[htb]{0.245\textwidth}
    \centering
    \captionsetup{justification=centering}
    \includegraphics[width=0.99\textwidth]{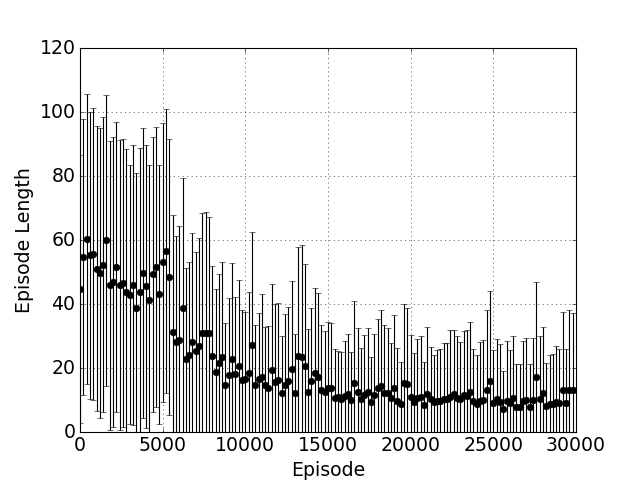}
    \captionsetup{width=0.95\textwidth}
    \caption{Meta-Controller with No Attention Mechanism on a Dynamic Environment}
    \label{fig:no-attention-dynamic}
    \end{subfigure}
    \begin{subfigure}[htb]{0.245\textwidth}
    \centering
    \captionsetup{justification=centering}
    \includegraphics[width=0.99\textwidth]{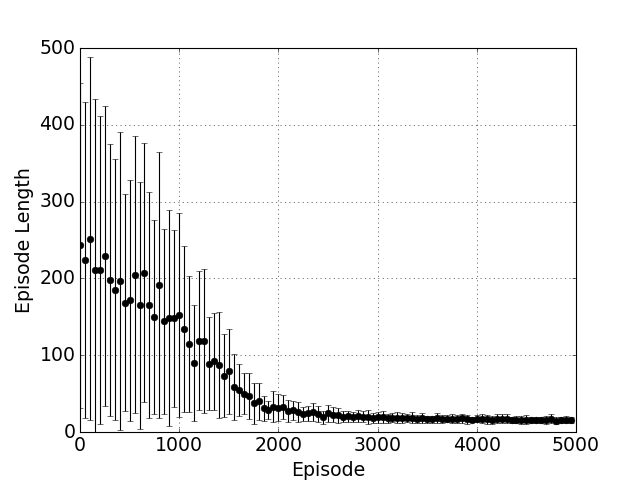}
    \captionsetup{width=0.95\textwidth}
    \caption{Meta-Controller with Partial State Decomposition}
    \label{fig:partial-fixed}
    \end{subfigure}
    \begin{subfigure}[htb]{0.245\textwidth}
    \centering
    \captionsetup{justification=centering}
    \includegraphics[width=0.99\textwidth]{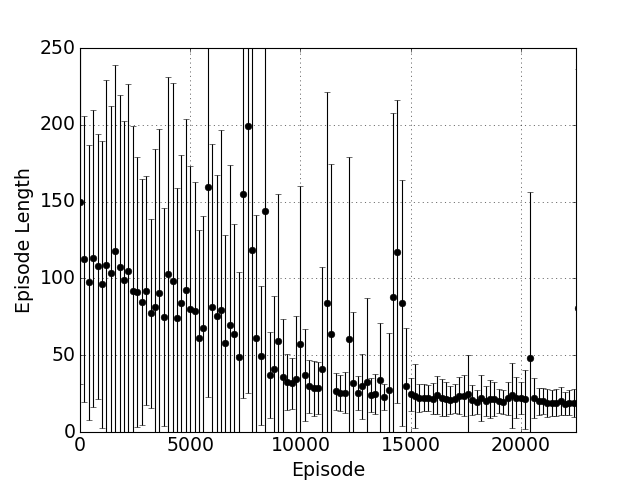}
    \captionsetup{width=0.95\textwidth}
    \caption{Meta-Controller with Constrained Attention Mechanism}
    \label{fig:constrained-fixed}
    \end{subfigure}
    \caption{Episode lengths over training episodes. The dots represent means over a variable number of episodes dependent on the total number of episodes displayed and the whiskers represent the variance. In the case of the meta-controller with no attention (a and b), it converges quickly on a fixed environment, supplying the agent with the target room and hence completing episodes in the minimum time. It takes longer to converge in the dynamic case. The meta-controller with attention plots (c and d) show the results on the effect of using attention to guide subgoal creation. For these experiments, the environment was kept fixed.}
\end{figure}

\subsection{Meta-Controller with Attention Mechanism}
Here, we show the results for the meta-controller with both the partial state decomposition and the constrained attention mechanism. In these experiments, we keep the environment fixed between episodes. Compared to Figure \ref{fig:no-attention-fixed}, in both cases, it takes longer to train the meta-controller to output subgoals leading to the optimal policy. One reason for this is that the meta-controller now has to control its attention in addition to creating subgoals. It is also operating in a partially observed setting and has to integrate information gleaned in past attentions into its hidden state. Note that the state space of the meta-controller is the set of combinations of attention window, target instruction, and hidden state of the LSTM unit. But since this is possible to learn, the underlying agent does not have to be trained on the entire $10$x$5$ image, but only on the $5$x$5$ attention size. Our approach may scale to even larger sized domains, where directly learning a policy on the original input image may be infeasible.

\section{Conclusion}
Our overall contribution is a framework that allows an agent to complete a task in a large environment given knowledge of how to do so in a smaller environment. Through the use of an attention mechanism, smaller networks are required, which are easier to train. With all three frameworks developed, the meta-controller learns the representation of the room colors and how that representation transfers to sub-instructions that lead the agent to the desired goal. Our results show that it is possible to scale a policy learned on a smaller environment by decomposing a large state space using an attention mechanism. Our eventual goal is to train the underlying agent in conjunction with the meta-controller and apply this framework to dynamic and complex environments. 

\bibliographystyle{unsrt}
\bibliography{references}

\end{document}